\documentclass[journal,twoside,web]{ieeecolor}
\usepackage{generic}
\usepackage{cite}
\usepackage{amsmath,amssymb,amsfonts}
\usepackage{graphicx}
\usepackage{placeins}
\usepackage{textcomp}
\usepackage{microtype}
\usepackage{xcolor}
\usepackage{etoolbox}
\usepackage{tabularx,booktabs,multirow}
\usepackage[font=small]{caption}
\usepackage{subcaption}
\usepackage{siunitx}
\usepackage[pdfdisplaydoctitle,colorlinks,allcolors=blue!40!black]{hyperref}

\providecommand{\refname}{References}
\patchcmd{\thebibliography}{\color{black}}{\color{black}\refname}{}{}
\patchcmd{\thebibliography}{\itemsep 0pt plus pt}{\itemsep 0pt}{}{}

\makeatletter
\def\ps@titlepagestyle{%
    \def\@oddhead{\hbox{}\scriptsize\textsf{\leftmark}\hfil\thepage}%
    \let\@evenhead\@oddhead
    \def\@oddfoot{}\def\@evenfoot{}%
}
\makeatother

\usepackage[sort&compress,nameinlink,noabbrev,capitalize]{cleveref}

\usepackage{tikz}
\usetikzlibrary{arrows,decorations.pathmorphing,decorations.pathreplacing,backgrounds,positioning,fit,matrix}
\usetikzlibrary{shapes,calc}
\usetikzlibrary{arrows.meta,positioning,calc,fit,backgrounds,shapes.symbols}
\usetikzlibrary{calc,fit,backgrounds}

\newcommand{\A}{\textbf{L}}

\newcommand{\x}{\textbf{x}}

\newcommand{\mymethod}{SoST}
\newcommand{\mymethodtc}{SoST+TC}
\newcommand{\CARMEN}{IBP2021}

\def\BibTeX{{\rm B\kern-.05em{\sc i\kern-.025em b}\kern-.08em
    T\kern-.1667em\lower.7ex\hbox{E}\kern-.125emX}}
\begin{document}
\bstctlcite{IEEEexample:BSTcontrol}
\title{Decoupling Time and Space: A Temporally Conditioned Refinement for EEG Source Imaging}
\author{Marco Morik, Jesse Palarus, Carmen Vidaurre, Klaus-Robert M\"uller, and Shinichi Nakajima
\thanks{This work has been submitted to the IEEE for possible publication. Copyright may be transferred without notice, after which this version may no longer be accessible.\\
This work was funded by the German Federal Ministry of Education and Research under the grant BIFOLD25B. Furthermore, K.-R. M\"uller was partly supported by the Institute of Information \& Communications Technology Planning \& Evaluation (IITP) grant funded by the Korea government (MSIT) (No. RS-2019-II190079, Artificial Intelligence Graduate School Program, Korea University) and grant funded by the Korea government (MSIT) (No. RS-2024-00457882, AI Research Hub Project), DFG and Hector Science Foundation.\\ Corresponding authors are M. Morik (m.morik@tu-berlin.de) and S. Nakajima (nakajima@tu-berlin.de)}
\thanks{
M. Morik, K.-R. M\"uller and S. Nakajima are with the Berlin Institute for the Foundations of Learning and Data, Germany and also with the Machine Learning Group, Technische Universität Berlin, Germany \\
    J. Palarus was with the Machine Learning Group, Technische Universität Berlin, Germany \\
 K.-R. M\"uller is additionally with the Department of Artificial Intelligence, Korea University, Seoul, Korea and also with the Max Planck Institut für Informatik, Saarbr\"ucken, Germany \\
 C. Vidaurre is with Basque Center on Cognition, Brain and Language, Mikeletegi Pasealekua, Spain, with Ikerbasque, Bizkaia, Spain, and with Department of Machine Learning, Technische Universität Berlin, Germany \\
 S. Nakajima is additionally with RIKEN AIP, Tokyo, Japan \\
}
}

\maketitle

\begin{abstract}
Electroencephalography (EEG) offers millisecond temporal resolution, but inferring underlying neural sources is a severely ill-posed spatial inverse problem. While deep learning has advanced spatial reconstruction, current architectures face a critical dilemma: frame-by-frame models discard vital temporal context, whereas full 4D spatiotemporal networks introduce an architectural trade-off between reconstruction accuracy and inference cost. We propose a novel two-stream framework that explicitly decouples global temporal representation learning from per-time-point spatial refinement. A Transformer-based Temporal Condition Encoder processes the entire EEG sequence via factorized spatiotemporal attention, retaining sensor-resolved features. A fixed inverse then maps these features into source-indexed conditioning for a per-timestep Source-Space Transformer or volumetric convolutional refiner. Extensive evaluations on realistic synthetic data demonstrate that this temporal prior dramatically improves spatial localization, outperforming classical and spatiotemporal baselines, particularly in high-noise and multi-source regimes. Training across diverse leadfields and explicit operator mismatches improves transfer to unseen head geometries and brings template-based reconstruction closer to subject-specific inversion. Furthermore, we apply the model trained only on synthetic EEG data to real-world EEG. A logistic regressor fit on source power differences in eyes-open, eyes-closed conditions successfully decodes age groups.
\end{abstract}


\section{Introduction}
\label{sec:introduction}

Understanding human brain dynamics requires imaging modalities with high temporal resolution and reliable spatial inference \cite{michel2004eeg, MichelBrunet2019}. Electroencephalography (EEG) and magnetoencephalography (MEG) provide millisecond-scale measurements of neural activity, enabling the study of fast cognitive processes \cite{gevins1995mapping,blankertz2008optimizing,haufe2008combining,blankertz2010berlin,blankertz2011single,burle2015spatial,he2018electrophysiological}. However, recovering the underlying neural sources from scalp sensor measurements remains fundamentally ill-posed \cite{baillet2001electromagnetic}.

In a typical clinical or research setting, the number of sensors $M$ is considerably smaller than the number of candidate sources $S$ whose activity is to be estimated. The forward model, $\mathbf{y}_t = \mathbf{L}\mathbf{x}_t + \boldsymbol{\epsilon}_t$, where $\mathbf{L}$ is the lead field matrix, is therefore highly underdetermined \cite{grech2008review}. The lead field varies substantially across subjects \cite{akalin2013effects,hansen2016data,harmening2026data}. Accurate reconstruction depends on forward-model fidelity, noise, and the imposed priors.

Classical distributed methods, such as minimum norm estimates (MNE) and eLORETA \cite{ELORETA}, introduce hand-crafted mathematical regularization (e.g., $L_2$ norms or spatial smoothness) \cite{hamalainen1994mne}. While computationally efficient, these approaches impose rigid assumptions that limit their ability to capture the diverse, sparse, and distributed patterns observed in true neural activity \cite{ding2008sparse}.

Deep learning methods provide a powerful data-driven alternative, learning complex priors from simulated data and directly mapping sensor measurements to source estimates \cite{delatolas2023eeg, sun2022deepsif, hecker2021convdip}. Although these end-to-end approaches have demonstrated improved performance \cite{zhang2024dct} with generic and personalized head models \cite{sun2023deep}, they typically incorporate the biophysical forward model only implicitly through their training data. Transfer under changed head models and noise therefore requires explicit evaluation.

Hybrid approaches combine physics-based initialization with neural refinement, for example by applying an eLORETA pseudo-inverse followed by volumetric convolutional networks \cite{morik2025enhancing}. While effective for spatial reconstruction, these methods process each time step independently, neglecting the highly autocorrelated nature of signals from biological neural networks.  Spatiotemporal architectures offer different trade-offs between reconstruction accuracy, spatial representation, temporal context, and inference cost. Approaches such as recurrent networks \cite{sun2022deepsif} are fast, but accuracy is limited, while full spatiotemporal convolutions that explicitly attempt to model time in source space incur severe computational costs. The size of this 4D spatiotemporal volume forces compromises, limiting either the spatial resolution of the source space or the extent of the temporal context.

To address these limitations, we propose a temporally conditioned framework (\mymethodtc) that explicitly decouples global temporal modeling from spatial reconstruction. Our proposed method distills a compact representation of the global temporal context by learning to solve the biophysical inverse problem strictly on simulated data. This design enables the efficient integration of long-range temporal priors to steer volumetric source reconstruction, while simultaneously learning a representation of temporal brain dynamics. Furthermore, we show that robust training with varying leadfields improves transfer across subject head geometries. Our operator-mixing protocol explicitly incorporates mismatches between data-generating and inversion leadfields during training, enabling template-based reconstruction to approach the accuracy of subject-specific inversion on held-out head geometries.

\textbf{Contributions:}
\begin{itemize}
    \item We propose a two-stream architecture featuring a factorized Transformer that efficiently extracts long-range temporal dependencies from full EEG sequences, explicitly handling temporal heterogeneity, and a 3D Transformer to refine physics-informed pseudo-inverse initializations.
    \item We demonstrate that this decoupled temporal prior improves spatial localization and robustness, outperforming classical and deep learning baselines in low-SNR and complex multi-source environments.
    \item We analyze the effect of training with a variety of head models, demonstrating the importance of selecting simulation settings that closely match the target application setting.
\end{itemize}

\section{Background and Related Work}

\subsection{Classical Approaches to Source Localization}
Traditional approaches to the EEG inverse problem rely on mathematical regularization to constrain the infinite solution space \cite{tarantola2005inverse}. Early parametric methods, such as equivalent current dipole (ECD) fitting, assume a small, \emph{a priori} fixed number of active focal regions and optimize their locations and orientations \cite{scherg1990fundamentals}. While effective for simple evoked potentials, these methods struggle with distributed brain networks and often converge to local minima \cite{baillet2001electromagnetic}.

Distributed inverse solutions bypass the need to specify the number of sources by discretizing the brain volume or cortical surface and jointly estimating amplitudes across all locations. To obtain a unique solution, these methods introduce spatial or energy-based priors. Minimum Norm Estimates (MNE) \cite{hamalainen1994mne} use an $\ell_2$-norm penalty to find the lowest-energy configuration and can favor superficial sources without depth weighting. Several extensions have been proposed to mitigate this limitation or improve localization properties, including dSPM \cite{dale2000dynamic}, LORETA \cite{pascual1994low}, sLORETA \cite{pascual2002standardized} and eLORETA \cite{ELORETA}.

When focal source estimates are desirable, sparsity can be promoted through $\ell_1$-norm regularization \cite{haufe2008combining,haufe2011large}. Many widely used linear methods estimate source activity independently at each time point, failing to leverage the complex, long-range oscillatory dynamics inherent to biological neural networks \cite{grech2008review}. Beamformers use the sensor covariance matrix across multiple observations to construct spatial filters \cite{vanveen1997lcmv,westner2022unified}. Some classical methods do include temporal information, e.g., TF-MxNE promotes sparse, non-stationary activity in the time-frequency domain \cite{gramfort2013time}, while Markov priors explicitly couple source activity over time \cite{hansen2017spatio}. Sparse Bayesian Learning (SBL) instead uses a hierarchical probabilistic model to estimate the source and noise hyperparameters from data \cite{wipf2009sparse,cai2021robust}. Spatiotemporal Bayesian formulations model temporal dependencies and structured noise \cite{hashemi2021efficient,hashemi2022joint}. Thus, unlike fixed linear inverse operators, these methods adapt aspects of the inverse solution to the statistics of the measured data.

\subsection{Deep Learning and Physics-Informed Approaches}
Driven by the availability of realistic large-scale brain simulations \cite{breakspear2017dynamic}, deep learning has emerged as a powerful alternative, replacing hand-crafted mathematical priors with data-driven priors. Early architectures focused on mapping instantaneous sensor topographies directly to 3D source spaces. For example, ConvDip \cite{hecker2021convdip} utilized convolutional neural networks to estimate distributed activity from single-timepoint EEG data, demonstrating improved robustness over classical methods.

More recently, hybrid \emph{physics-informed} architectures have bridged the gap between classical theory and deep learning. For the sparse setting, recent work augments iterative Bayesian inference by learning structured corrections to the underlying update procedure \cite{morik2026structure}. For the dense setting, approaches such as 3D-PIUNet \cite{morik2025enhancing} compute an initial, physics-based solution via an eLORETA pseudo-inverse and subsequently refine this volumetric estimate using a 3D CNN. These approaches anchor the network in the biophysics of the forward model. However, these spatial refiners operate frame-by-frame, discarding the rich temporal structure of the EEG signal.

\subsection{Spatiotemporal and Sequence Modeling}
To address the limitations of instantaneous inference, another line of work integrates temporal modeling. Recurrent neural networks support neuroimaging decoding\cite{thomas2019analyzing} and model brain dynamics over time~\cite{sun2022deepsif,jiao2022graph}. Recently, transformer encoders have also been applied to EEG source reconstruction~\cite{zheng2023transformeresi}. Closely related, SSINet combines spatial residual features and bidirectional LSTM features through Transformer-based fusion~\cite{shi2025deep}. However, treating high-resolution source localization as a joint 4D mapping problem (3D space + 1D time) is computationally demanding. Therefore, these architectures offer different trade-offs between temporal context, spatial representation, and inference cost.

Concurrently, within self-supervised learning (SSL), EEG foundation models such as ALFEE \cite{xiong2025alfee}, NeurIPT \cite{Fang2025NeurIPT}, CEReBrO \cite{dimofte2025cerebro} and CBraMod \cite{cbramod} use factorized attention mechanisms, explicitly separating temporal self-attention from inter-channel spatial attention, to efficiently process long, multichannel recordings. While these models demonstrate that decoupling time and space is highly effective for sequence modeling, they are designed strictly for representation learning, not for inverting the ill-posed biophysical forward model.

\begin{figure*}
    \centering
    \input{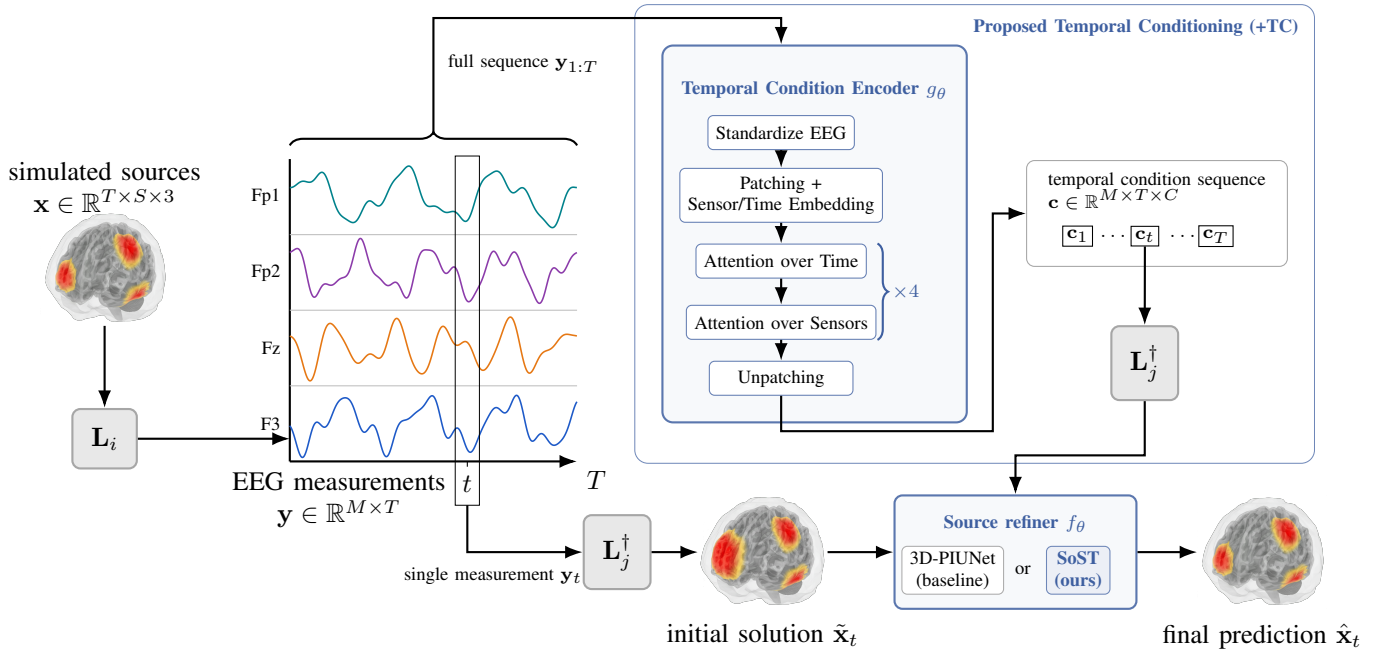}
    \vspace{-0.3cm}
    \caption{Temporal conditioning for source refinement. The novel Temporal Conditioning (+TC) path (highlighted in blue) applies the Temporal Condition Encoder $g_\theta$ to the full EEG sequence, alternates temporal and sensor attention, and unpatches features to an aligned sequence of condition vectors $\mathbf{c}$. At each time point, the subject-specific eLORETA-inverse $\mathbf{L}_j^{\dagger}$ transports the EEG slice and each condition feature into source space. The refiner $f_\theta$ fuses these two source-space inputs to predict the final sources $\hat{\mathbf{x}}_t$.}
    \label{fig:architecture_overview}
\end{figure*}

\section{Methods}
\label{sec:method}

To combine spatial reconstruction and long-range temporal modeling, we propose a novel two-stream framework. As illustrated in \cref{fig:architecture_overview}, the architecture consists of a global Temporal Condition Encoder ($g_\theta$) and a per-time-point Source Refiner ($f_\theta$). For any given timestep $t$, the final source prediction $\hat{\mathbf{x}}_t$ is formalized as:
\begin{equation}
    \hat{\mathbf{x}}_t = a_t f_\theta \left( \frac{\tilde{\mathbf{x}}_t}{a_t},\, \frac{\mathbf{c}'_t}{d_t} \right),
    \label{eq:prediction}
\end{equation}
where $\tilde{\mathbf{x}}_t=\mathbf{L}_j^\dagger\mathbf{y}_t$ is the initial source estimate, $\mathbf{c}'_t=\mathbf{L}_j^\dagger g_\theta(\bar{\mathbf{y}})_t$ the conditioning vector transported into source space, and $\mathbf{L}_j^\dagger\in\mathbb{R}^{3S\times M}$ is the pre-computed eLORETA inverse \cite{ELORETA} used for inversion. To provide well behaved inputs to the neural network, at each time point, the branches are independently scaled by $a_t=\max|\tilde{\mathbf{x}}_t|+10^{-8}$ and $d_t=\max|\mathbf{c}'_t|+10^{-8}$ over their real source entries and all feature/orientation components. This formulation ensures that both the primary input and the temporal conditioning signal are physically anchored in the 3D source space before non-linear neural refinement occurs. In the following, we explain each of the components in detail.

\subsection{Factorized Temporal Condition Encoder}
The Temporal Condition Encoder ($g_\theta$) distills the full EEG recording into temporally aware, sensor-resolved features. First the EEG signal is normalized across all sensors and the full sequence. A shared 1D convolution with kernel length 16, stride 2, padding 7, and GELU embeds overlapping windows into hidden width $H=96$. For $T=512$, this yields $P=256$ patches and a token array of shape $M\times P\times H$.

To preserve the biophysical topology of the recording, we inject 3D Cartesian spatial encodings via independent 1D sinusoidal encodings to the known $(x, y, z)$ coordinates of each electrode. This encodes electrode positions and permits coordinate-aware inputs \cite{el2026reve}. Additionally, we use a 1D sinusoidal encoding depending on the time, providing each token with temporal information. To have uniform token counts within a batch, variable sensor/source counts are padded, masks exclude padded entries from attention, normalization statistics, and source losses.

\textbf{Factorized Spatiotemporal Attention:} A Transformer \cite{vaswani2017attention} operating on the flattened $M \times P$ grid requires $\mathcal{O}((MP)^2)$ pairwise interactions. Inspired by efficient video architectures \cite{gberta_2021_ICML} and EEG sequence models \cite{dimofte2025cerebro}, we employ factorized attention.
The core network consists of $N=8$ transformer blocks that alternately perform self-attention over time and sensor dimensions, $4$ blocks each, with each block having eight heads, feed-forward width 384 and dropout 0.1:
\begin{enumerate}
    \item \textit{Time-Level Attention:} Self-attention across the $P$ temporal patches independently for each sensor captures sequence-wide context.
    \item \textit{Sensor-Level Attention:} Self-attention is applied across the $M$ sensors independently for each temporal patch, allowing spatial integration of instantaneous topographies.
\end{enumerate}
This factorization reduces the attention interactions per block to $\mathcal{O}(PM^2 + P^2M)$. After the final block, the patch-based representation is upsampled via a linear projection to match the original temporal resolution $H\to 2C$, yielding the raw condition matrix $\mathbf{c}\in\mathbb{R}^{M\times T\times C}$ with output width $C=32$, maintaining both the sensor and temporal dimensions of the input.

\subsection{Source-Space Transformer (\mymethod)}
The source refiner ($f_\theta$) operates independently at each time point, using the physics-informed inverse initialization $\tilde{\mathbf{x}}_t=\mathbf{L}_j^\dagger\mathbf{y}_t\in\mathbb{R}^{3S}$ as input to refine the source estimate.

While prior hybrid methods rely on a 3D U-Net \cite{morik2025enhancing}, their convolutional core requires an equally spaced voxel grid. For irregular surface and volumetric source locations, we propose the Source-Space Transformer (\mymethod) based on a DiT-style Transformer backbone adapted for supervised regression \cite{DiT}. \mymethod{} uses one token per source location with 3D sinusoidal positional encodings and applies global self-attention across all source locations. The network consists of six transformer blocks of width 384 with six attention heads and feed-forward expansion factor of four. Stochastic depth \cite{huang2016deep} increases from zero to 0.1. Each token predicts three dipole components.

\subsection{Physics-Informed Temporal Conditioning} \label{sec:conditioning}
\textbf{Source-Space Concatenation} uses the sensor-resolved temporal features $\mathbf{c}_t\in\mathbb{R}^{M\times C}$, applying the inverse operator independently to each latent feature, thereby producing source-indexed conditioning:
\begin{equation}
    \mathbf{c}'_t = \mathbf{L}_j^{\dagger} \mathbf{c}_t \quad \in \mathbb{R}^{3S \times C}.
\end{equation}
Applying the fixed inverse to each latent feature supplies a source-indexed inductive bias and associates the $C$-dimensional temporal feature vector of a sensor with the specific 3D source locations and orientations derived from the leadfield.

The scaled spatial estimate $\tilde{\mathbf{x}}_t$ is then concatenated with the scaled source-space condition $\mathbf{c}'_t$ along the feature dimension, yielding a per-source feature array of size $S\times(3+3C)=S\times99$. By providing the temporal prior directly as a spatially-resolved input feature, \mymethodtc{} can dynamically weigh the temporal context at each source location.

\subsection{Biophysical Data Generation Pipeline} \label{sec:data_gen}
\begin{table}[t]
\centering
\caption{Lead-field pools. $M$/$S$ denote sensor/source-location counts, the final column counts training/evaluation lead fields. Surface and volume spaces use \texttt{oct4} and 14-mm grids, respectively.}
\label{tab:head_model_pools}
\scriptsize
\setlength{\tabcolsep}{3pt}
\begin{tabular}{@{}llrrr@{}}
\toprule
\textbf{Space} & \textbf{Head model} & \textbf{$M$} & \textbf{$S$} & \textbf{\# (train/eval)} \\
\midrule
\multirow{3}{*}{Surface}
 & LEMON~\cite{babayan2019mind} & 61 & 516 & 127/10 \\
 & \CARMEN~\cite{nierhaus2021immediate} & 62 & 516 & 11/5 \\
 & \texttt{fsaverage}~\cite{fischl1999high} & 16-62 & 516 & 2-3/1 \\
\midrule
\multirow{2}{*}{Volume}
 & \texttt{fsaverage}~\cite{fischl1999high} & 16-62 & 599 & 3/1 \\
 & \CARMEN~\cite{nierhaus2021immediate} & 62 & 461-641 & 11/5 \\
\bottomrule
\end{tabular}
\end{table}
Our source models are trained entirely on synthetic EEG. Therefore, the synthetic data-generating process defines the prior learned by the models, with the richness of the simulated training distribution determining its inductive bias. Selected source locations remain fixed within each trial, imposing a prior of temporally stable source support.
The generation pipeline explicitly models spatiotemporal neural dynamics and samples a range of source extents, oscillatory time courses, noise levels, and head geometries.

Each training sample ($\mathbf{Y} \in \mathbb{R}^{M \times T}$) is generated through a four-stage process:

\textbf{1. Spatial Configuration:} The number $K$ of active source patches is sampled uniformly from $\{1,\ldots,20\}$. Each Gaussian patch is centered at a uniformly sampled location on the discretized brain and has a spatial standard deviation $\sigma_k\sim\mathcal{U}(5,40)$\,mm. Its spatial weights have unit $\ell_2$ norm. A normalized isotropic Gaussian vector supplies one orientation per patch, yielding $\mathbf{S}^{(k)}\in\mathbb{R}^{3S}$.

\textbf{2. Temporal Dynamics:} To reflect canonical EEG rhythms, each source is assigned a band-limited time course $\mathbf{s}_k\in\mathbb{R}^T$ generated by forward-backward bandpass filtering Gaussian white noise and normalizing it to unit $\ell_2$ norm. The band is sampled uniformly from delta (\qtyrange{0.5}{4}{\hertz}), theta (\qtyrange{4}{8}{\hertz}), alpha (\qtyrange{8}{13}{\hertz}), mu (\qtyrange{8}{12}{\hertz}), and beta (\qtyrange{12}{30}{\hertz}). With probability 0.5, a time course is modulated by smooth Gaussian ERD/ERS-inspired amplitude envelopes \cite{pfurtscheller1999erd}. The full source activity is $\mathbf{X}=\sum_{k=1}^K\mathbf{S}^{(k)}\otimes\mathbf{s}_k$.

\textbf{3. Forward Projection and Head-Model Diversity:}
Because source reconstruction is highly sensitive to the forward model \cite{stahlhut2011simultaneous}, we train across a pool of lead fields to increase robustness to head-model variability. For each simulated sample, we first sample a lead field $\mathbf{L}_i$ from the pool in \cref{tab:head_model_pools}. Lead fields for individual subjects are computed using three-layer BEMs \cite{fuchs2001bem} implemented in MNE-Python \cite{GramfortEtAl2013a}. We additionally introduce a conductivity mismatch between training and evaluation for \CARMEN, using brain-skull-scalp conductivities of $(0.300,0.006,0.300)\,\mathrm{S\,m^{-1}}$ during training and for the inverses, but
$(0.332,0.0113,0.332)\,\mathrm{S\,m^{-1}}$ during data generation for evaluation \cite{wolters2010combined}. Sensor-space measurements are generated as $\mathbf{Y}=\mathbf{L}_i\mathbf{X}$.

For \textbf{operator-mix (OM)} training, measurements use $\mathbf{L}_i$, while both inverse-input and conditioning branches use $\mathbf{L}_j^\dagger$. The curriculum begins with matched operators ($j=i$) and ramps mismatches linearly over 10-30\% of training, reaching 50\% $(s,t(s))$, 20\% $(s,s)$, 10\% $(s,s')$, and 20\% $(t,t)$. Here, $s$ is a subject, $t(s)$ its dataset template, $s'\neq s$ another subject from the same dataset, and $t$ a sampled template. Subject-forward draws balance eligible datasets, then subjects, the template branch remains at 20\%. For stored template-to-subject interpolation $\mathbf{G}_s$, generation uses $\mathbf{L}_s(\mathbf{G}_s\otimes\mathbf{I}_3)$, a subject inverse $j$ uses target $(\mathbf{G}_j\otimes\mathbf{I}_3)\mathbf{X}$. A template inverse retains $\mathbf{X}$. For our free orientation setting, the componentwise interpolation introduces no rotation and assumes compatible Cartesian orientation bases.

\textbf{4. Noise:} To reflect the severe corruption present in real clinical recordings, we inject two distinct noise components:
\begin{enumerate}
    \item \textit{Biological Background Noise:} Oriented pink noise with power spectral density proportional to $1/f$ is projected from $50$ to $500$ random source locations.
    \item \textit{Sensor Noise:} Zero-mean white Gaussian noise is added directly to the projected sensor measurements.
\end{enumerate}
For training, nominal SNR is uniform over $[-10,30]$\,dB. Using sample variance $P_y$ over real sensors and time, the target noise variance is $P_n=P_y10^{-\mathrm{SNR}/10}$. Projected pink and sensor noise are scaled independently to variances $\rho P_n$ and $(1-\rho)P_n$, with $\rho\sim\mathcal{U}(0.80,0.95)$.

This simulation provides an effectively unlimited supply of labeled pairs $(\mathbf{Y},\mathbf{X})$, enabling the network to learn robust spatiotemporal inversion under varied noise and lead fields.

\section{Experimental Setup}
\label{sec:exp_setup}
\subsection{Evaluation Metrics}
We report five quantities in three metric families, following \cite{morik2025enhancing}. Let $\mathbf{x}_{ts},\hat{\mathbf{x}}_{ts}\in\mathbb{R}^3$ be target and predicted dipoles at real source locations, with magnitudes $a_{ts}=\|\mathbf{x}_{ts}\|_2$ and $\hat a_{ts}=\|\hat{\mathbf{x}}_{ts}\|_2$.

\textbf{NMAE:} Signed-vector amplitude error $\sum_{t,s,o}|\hat x_{tso}-x_{tso}|/\sum_{t,s,o}|x_{tso}|$, normalized by the $L_1$-norm of the ground truth to account for global signal scaling.

\textbf{NEMD:} Earth Mover's Distance \cite{Rubner2000} transports normalized magnitudes. For predicted and target probability vectors $\mathbf{p},\mathbf{q}$, we use $W_D(\mathbf{p},\mathbf{q})/W_D(\mathbf{u},\mathbf{q})$, where $\mathbf{u}$ is uniform on the same support and $W_D$ is optimal transport cost. Spatial costs are Euclidean source distances, temporal costs are sample distances $|t-t'|$. \emph{NEMD} averages independently normalized per-time-point spatial costs, \emph{Spatial NEMD} transports time-averaged magnitudes $\frac1T\sum_t a_{ts}$, \emph{Temporal NEMD} transports source-averaged magnitudes $\frac1S\sum_s a_{ts}$.

\textbf{Weighted Cosine Error:} We measure orientation fidelity using the cosine-based error $1-\sum_{t,s}a_{ts}\cos_\epsilon(\hat{\mathbf{x}}_{ts},\mathbf{x}_{ts})/\sum_{t,s}a_{ts}$, with each cosine norm floored at $10^{-8}$. NMAE and this error require nonzero target mass.

\subsection{Training Dynamics and Baseline Comparisons}
We benchmarked \mymethodtc{} against \mymethod{} without temporal conditioning, DeepSIF, an LSTM-based source-imaging model \cite{sun2022deepsif}, eLORETA \cite{ELORETA}, and a free-orientation adapted version of Deep CB, a learned SBL approach \cite{morik2026structure}. Additionally, the volume comparison evaluates TC/no-TC variants of \mymethod{} and 3D-PIUNet \cite{morik2025enhancing}.

The refiner is trained using an $\ell_1$ reconstruction loss in the normalized inverse-input space,
\[
\mathcal{L}_{\mathrm{ref}}
=
\frac{1}{B}
\sum_{b=1}^{B}
\sum_{t}
\sum_{s\in\mathcal{S}_b}
\left\|
\frac{
\hat{\mathbf{x}}_{bts}-\mathbf{x}_{bts}
}{
\alpha_{bt} |\mathcal{S}_b|
}
\right\|_1 ,
\]
Here, $\mathcal{S}_b$ contains all valid, non-padded source locations of sample $b$, including locations with zero target activity and $\alpha_{bt}$ is the inverse-input scaling factor, ensuring that the loss is computed in the normalized source space.
Training uses AdamW \cite{AdamW}, zero weight decay, 300 epochs, gradient clipping, and cosine learning-rate decay to $10^{-6}$. SoST uses initial learning rate $5\times10^{-4}$, batch size 16 and two-batch gradient accumulation, PIUNet uses $2\times10^{-4}$ and batch size 32. Each 512-point, 256\,Hz sequence supplies 32 loss time points: with equal probability, a uniformly located contiguous window or uniformly sampled indices without replacement. The encoder always sees the full sequence, DeepSIF and Deep CB use all time points. Each epoch contains 10,000 samples, where 1,000 are replaced each epoch on a rolling basis to increase training diversity.

\section{Experiments and Results} \label{sec:results}
In the following, we present our experimental results. We first evaluate synthetic reconstruction accuracy and the added cost of TC, then examine robustness, dependence on conditioning, and transfer across head-model pools. Finally, LEMON age-group decoding assesses downstream utility on real EEG.

\begin{figure*}[t]
    \centering
    \includegraphics[width=\linewidth]{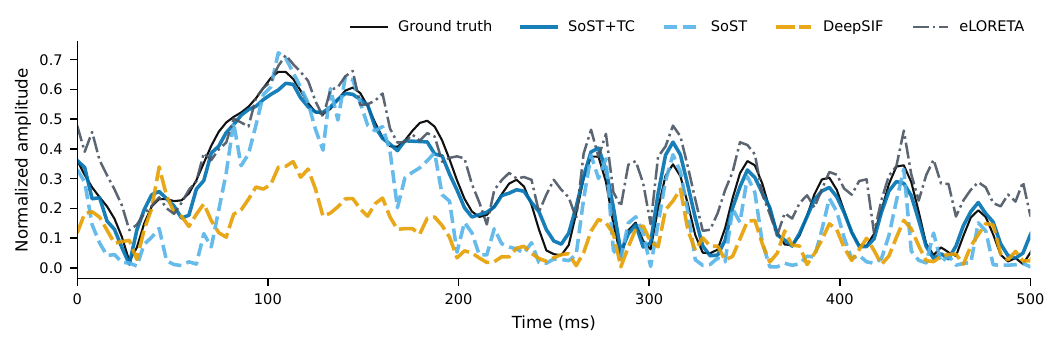}
     \caption{Reconstructed magnitudes at one active patch center on an unseen \CARMEN{} head at \qty{5}{\decibel} SNR. \mymethodtc{} (ours), \mymethod{}, DeepSIF, and eLORETA process the same sequence. The displayed 0-500\,ms window is taken from the 512-sample, 256\,Hz record.
    }
    \label{fig:temporal_amplitudes}
\end{figure*}

\subsection{Qualitative and Quantitative Performance}
Source reconstruction must preserve both the spatial distribution of activity and its amplitude over time. The temporal trajectory in \cref{fig:temporal_amplitudes} illustrates this aspect at one active patch center. \mymethodtc{} follows the displayed ground-truth magnitude trajectory more closely than \mymethod{}. This example illustrates the potential of temporal context to stabilize reconstruction in the noisy, ill-posed inverse problem.

\begin{table*}[t]
\centering
\caption{Aggregate reconstruction errors (lower is better) and computational cost. Evaluation uses 200 simulated sequences per head (512 samples at 256\,Hz), with SNR, active patch count, and patch standard deviation sampled uniformly from $[-10,30]$\,dB, $\{1,\ldots,20\}$, and $[5,40]$\,mm, respectively. Within each surface seed, errors are averaged over ten held-out LEMON and five held-out \CARMEN{} heads separately, then equally across datasets, taking the mean over three training seeds. Volume rows average five held-out \CARMEN{} heads for one training seed. Runtime protocol: FP32, batch size one, 512 time points per forward call, 50 timed runs on an NVIDIA A100 80GB.}
\label{tab:accuracy_runtime}
\scriptsize
\setlength{\tabcolsep}{3.5pt}
\resizebox{\textwidth}{!}{\begin{tabular}{@{}lrrrrrrr@{}}
\toprule
Method & NMAE & NEMD & Spatial NEMD & Temporal NEMD & Cosine Error & Params. (M) & Latency (ms) \\
\midrule
\multicolumn{8}{@{}l}{\emph{Surface source space}} \\
SoST & 0.824 & 1.023 & 0.804 & 0.572 & 0.354 & 16.5 & 489.8 \\
SoST+TC & 0.645 & 0.701 & 0.652 & 0.255 & 0.219 & 17.4 & 504.8 \\
SoST+TC+OM & \textbf{0.577} & \textbf{0.641} & \textbf{0.593} & \textbf{0.232} & \textbf{0.185} & 17.4 & 505.9 \\
Deep CB~\cite{morik2026structure} & 0.889 & 0.844 & 0.819 & 0.404 & 0.380 & $<0.1$ & 50.2 \\
DeepSIF~\cite{sun2022deepsif} & 0.860 & 0.937 & 0.873 & 0.728 & 0.399 & 7.4 & 19.3 \\
eLORETA~\cite{ELORETA} & 1.941 & 0.902 & 0.866 & 0.493 & 0.498 & 0.0 & 0.4 \\
\midrule
\multicolumn{8}{@{}l}{\emph{Volume source space}} \\
SoST & 0.798 & 1.028 & 0.750 & 0.627 & 0.343 & 16.5 & 430.0 \\
SoST+TC & \textbf{0.577} & \textbf{0.657} & \textbf{0.617} & \textbf{0.256} & \textbf{0.186} & 17.4 & 444.3 \\
3D-PIUNet~\cite{morik2025enhancing} & 0.810 & 1.020 & 0.765 & 0.643 & 0.351 & 19.1 & 126.7 \\
3D-PIUNet+TC & 0.666 & 0.800 & 0.803 & 0.320 & 0.234 & 20.0 & 150.1 \\
eLORETA~\cite{ELORETA} & 1.709 & 0.908 & 0.852 & 0.476 & 0.433 & 0.0 & 0.4 \\
\bottomrule
\end{tabular}
}
\end{table*}
To assess aggregate reconstruction accuracy and computational cost, we provide quantitative analysis in \cref{tab:accuracy_runtime}. \mymethodtc{} outperforms the evaluated learned and classical baselines in reconstruction accuracy, at the cost of greater inference time. However, the cost of temporal conditioning is modest, on the surface source space, it reduces \mymethod{}'s NMAE by 21.7\% while increasing measured inference latency by only 3.1\%. Incorporating sequence context therefore yields a substantial accuracy improvement for a small additional cost over the unconditioned refiner.
The volumetric experiments further distinguish the contribution of the reconstruction architecture from that of temporal conditioning. On the volume source space, the transformer architecture \mymethod{} performs comparably to the convolutional 3D-PIUNet~\cite{morik2025enhancing}. Adding temporal conditioning also improves 3D-PIUNet, its benefit is therefore not restricted to the transformer architecture, although the improvement is greater for \mymethod{}. Together, these results support temporal conditioning as a useful addition to both evaluated reconstruction backbones, rather than an advantage specific to one architecture.

Operator-mismatch training provides a further improvement for \mymethodtc{}. Although introduced primarily to address head-model generalization (see \cref{sec:headmodelgeneralization}), it also benefits the reconstruction results summarized here.

\subsection{Robustness to Noise and Complex Source Settings}
\begin{figure}[t]
    \centering
    \includegraphics[width=\linewidth]{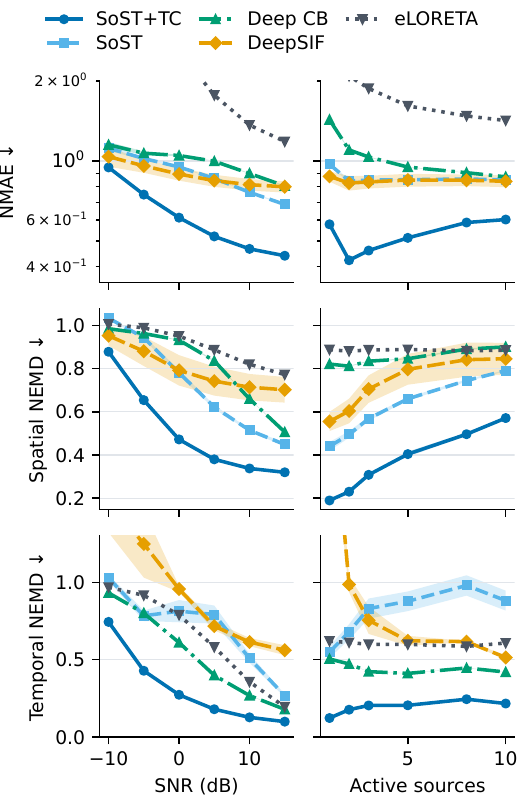}
    \caption{Robustness to sensor noise and source multiplicity. Rows show NMAE, spatial NEMD, and temporal NEMD. The SNR sweep (left) draws source counts uniformly from 1-10, the source-count sweep (right) fixes SNR at 5\,dB. Each point first averages 15 equally weighted held-out heads (ten LEMON, five \CARMEN{}) within each seed, then reports the mean $\pm$ sample SD over three training seeds. Our proposed \mymethodtc{} retains lower errors across both the evaluated noise and source-count settings.}
    \label{fig:snr_triptych}
    \label{fig:nsources_triptych}
\end{figure}

A primary failure mode of EEG inverse solvers is the high susceptibility to sensor and environmental noise, as well as a rise in complexity depending on the number of simultaneously active sources. In Fig. \ref{fig:snr_triptych}, we evaluated reconstruction quality across Signal-to-Noise Ratios (SNR) ranging from \qty{-10}{\decibel} (highly corrupted) to \qty{15}{\decibel} (moderate noise), as well as the number of sources ranging from $K=1$ to $K=10$.

At an SNR of \qty{-10}{\decibel}, \mymethodtc{} already detects some spatial structure. Across the evaluated noise levels, \mymethodtc{} reduces reconstruction errors relative to its time-agnostic variant \mymethod{}. This is consistent with temporal context providing complementary information when instantaneous sensor data are noise-dominated.

As source count increases, its spatial marginal NEMD rises, but remains below the compared baselines. It learns to disentangle spatially overlapping source contributions by exploiting their distinct temporal and oscillatory signatures. Temporal NEMD remains comparatively low, showing the advantage of TC in encoding temporal activity also beyond single-source examples.

\subsection{Role of conditioning vector}
\begin{figure}[t]
    \centering
    \includegraphics[width=\columnwidth]{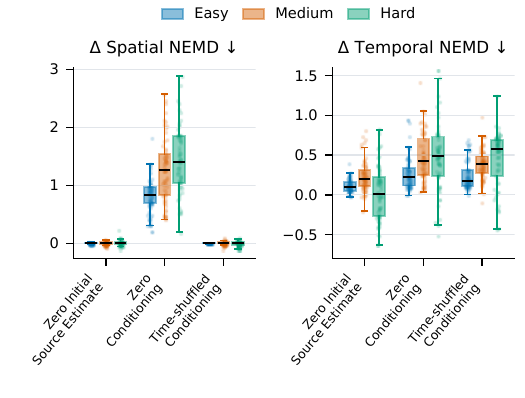}
    \caption{Conditioning interventions on one held-out \CARMEN{} head, with 64 paired simulations per difficulty. Left: spatial NEMD, right: temporal NEMD. Positive deltas indicate worse errors than the intact input. Boxes show medians and interquartile range, and points show individual simulations. Interventions are zero initial source estimate input, zero condition, and time-shuffled condition. Easy/medium/hard use 15/5/$-5$\,dB SNR and 1/1-5/5-10 active sources.}
    \label{fig:conditioning_robustness}
\end{figure}
To examine how the trained model uses conditioning, we intervene on its inputs for one held-out \CARMEN{} head. Figure~\ref{fig:conditioning_robustness} shows strong dependence on conditioning: zeroing the condition increases errors more than zeroing the initial source estimation branch. Note that spatial NEMD is invariant to permuting prediction time indices, and temporal NEMD measures total activation over time, not source-specific timing. We observe that time-shuffled conditioning primarily increases temporal NEMD, the little change in spatial NEMD shows that spatial structures are indeed reconstructed, however at the wrong points in time. Overall, this demonstrates that \mymethodtc{} strongly relies on the time-step conditioning vector for source reconstruction.

\subsection{Head Model Generalization}
\label{sec:headmodelgeneralization}
\begin{figure*}[t]
    \centering
        \includegraphics[width=\textwidth]{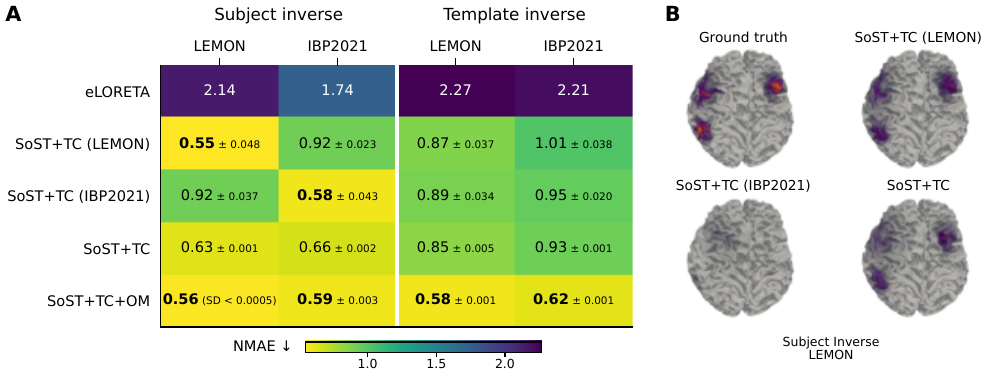}%
    \caption{Head-model generalization. \textbf{A:} NMAE (lower is better) averages ten held-out LEMON or five held-out \CARMEN{} heads in the respective columns, then reports seed mean $\pm$ sample SD over three seeds. Parentheses identify single-pool training. Specialists transfer less accurately across datasets, mixed training reduces transfer loss. Additionally, Operator Mixing improves the template inverse performance. \textbf{B:} Maps show full-epoch source-vector RMS divided by one common ground-truth maximum. The mismatched specialist underestimates activity and misses one of three active regions.}
    \label{fig:head_generalization}
\end{figure*}
We next examine whether reconstruction accuracy is maintained when the head geometries encountered at inference differ from those used during training. \Cref{fig:head_generalization} distinguishes two aspects of this problem: the coverage of head geometries in the training pool, and the consistency between the data-generating leadfield and the leadfield used to construct the inverse operator. We compare \mymethodtc{} trained only on LEMON geometries, only on \CARMEN{} geometries, or on both pools, and evaluate each model using subject-specific and template inverses.

Training on one geometry pool leads to worse performance on other unseen head models, while combined training reduces cross-dataset transfer loss. Although pooled-model errors remain higher than those of the corresponding within-pool specialists, they are substantially lower than the errors obtained by applying a specialist to the other dataset's heads. Pooling therefore reduces the dependence on a single training geometry pool while retaining good reconstruction accuracy on held-out heads from both datasets. Panel~B illustrates a spatial consequence of this transfer loss. For the displayed LEMON example, the \CARMEN{}-trained specialist strongly attenuates the reconstructed activity and fails to recover one of the three active regions. In contrast, the LEMON-trained specialist and the pooled model retain all three regions.

Broader geometry coverage alone does not eliminate sensitivity to the inverse operator. Under matched-operator training, the data-generating leadfield $L_i$ and the leadfield used to construct the inverse, $L_j$, are equal. Replacing the subject-specific inverse with an fsaverage template inverse introduces a mismatch not explicitly represented by this training setup, increasing the error. Our operator-mixing protocol explicitly includes $L_i \neq L_j$ during training. It reduces template-inverse NMAE, while it also improves subject-specific inversion.

The main practical consequence is a substantially smaller penalty for template-based inversion. Thus, explicitly training for operator mismatch substantially reduces dependence on a subject-specific inverse operator.


\subsection{Real-World Age Decoding on LEMON}
\label{sec:lemon_real}

\begin{figure*}[t]
    \centering
    \includegraphics[width=\textwidth]{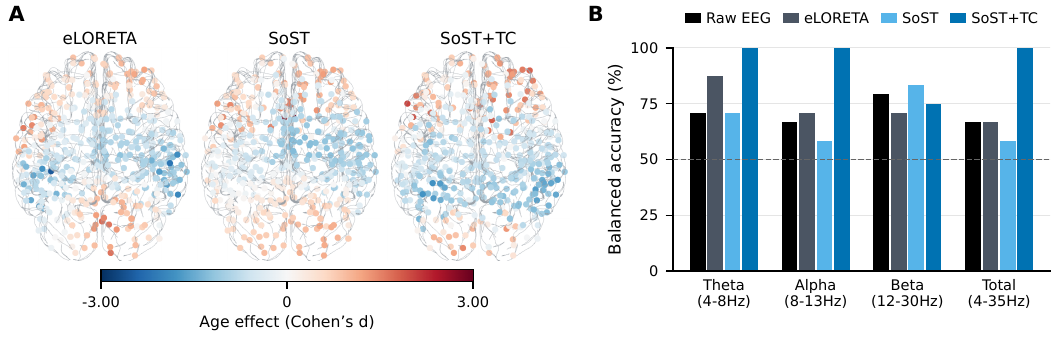}
    \caption{LEMON age-group decoding on 10 unseen test subjects, whose geometries were held out from source-model training. \textbf{A:} Descriptive older-minus-younger Cohen's $d$ maps from these ten participants, using pooled within-age SD within the 4-35\,Hz frequency band. \textbf{B:} Test balanced accuracy of separate logistic classifiers trained on 115 participants using one indicated band or total-power reactivity (516 source features, 61 raw-EEG features where missing electrodes use training-fold mean imputation). \mymethodtc{} shows great age-group decoding.}
    \label{fig:lemon_age_reactivity}
\end{figure*}
Neuroimaging-based brain-age estimates are being investigated as biomarkers of brain health~\cite{hofmann2022towards}. Age-related EEG differences span multiple frequency bands. Previous work~\cite{barry2017eeg} reported age differences across delta, theta, alpha, and beta, together with preserved overall eyes-open/eyes-closed reactivity but more focal topographic changes in older adults. Source-space analyses have also reported age-related shifts in the spatial distribution of resting power~\cite{perinelli2022power}. These observations motivate examining spatially resolved reactivity across multiple frequency ranges rather than focusing exclusively on alpha power.

Resting-state electrophysiology has also been used for age prediction. In a benchmark that included LEMON, Engemann et al.~\cite{engemann2022reusable} pooled eyes-open (EO) and eyes-closed (EC) recordings before feature extraction and identified their relationship as a question for further investigation in brain-age modeling. Here, we evaluate whether source-space EO-EC reactivity supports age-group decoding (20-35 versus 55-80 years) in LEMON~\cite{babayan2019mind}. Paired recordings and required metadata yielded 115 training participants. For testing, we selected the ten participants (six younger, four older) whose geometries were excluded from synthetic source-model training.

Features were unadjusted EO-EC differences in $\log_{10}$ subject-mean power, with 516 template-source features per band.  Thus, reactivity represents the logarithm of the EO-to-EC power ratio rather than absolute power in either condition. Welch power used Hann windows of up to two seconds, 50\% overlap, and summed orientation-component powers. $L_2$-regularized logistic regression used class balancing, training-only feature standardization, and a 0.5 older-class probability threshold. Four-fold development CV selected the logistic-regularization parameter $C\in\{10^{-3},10^{-2},\ldots,10^2\}$ by balanced accuracy. Final classifiers were fitted on all 115 participants. The source models were trained only on synthetic EEG.

Figure~\ref{fig:lemon_age_reactivity}A shows spatially heterogeneous age contrasts, with both positive and negative effects whose distribution and magnitude depend on the reconstruction method. The maps describe age differences in state-dependent modulation, not simply reduced or increased resting activity. Compared with eLORETA and SoST, \mymethodtc{} displays a different spatial balance of these effects, including a prominent bilateral negative-effect pattern.

The frequency profile of age decoding also depends on the representation (Fig.~\ref{fig:lemon_age_reactivity}B). Scalp-EEG features perform best in beta, whereas eLORETA features perform best in theta. \mymethodtc{} achieved 100\% balanced accuracy for theta, alpha, and total 4-35\,Hz power, and 75\% for beta. Relative to SoST, adding the TC component improves performance in theta, alpha, and total power on this test set, but not in beta, where SoST and scalp EEG perform better. Thus, the result is not a uniform advantage across frequency bands, rather, the source representation changes which spectral features are most useful for this decoding task.

These findings demonstrate that source models trained on synthetic EEG can provide discriminative features for real recordings with unseen geometries. The spatial maps offer candidate patterns for follow-up investigation of age-associated reactivity. The ten-participant test set provides limited precision for estimating generalization performance, and the effect-size maps are descriptive. Larger, independently evaluated samples are needed to establish reproducible spatial signatures or broad decoding advantages. Neither these maps nor subject-mean power decoding directly validates anatomical localization or temporal stability.


\section{Discussion and Conclusion} \label{sec:conclusion}
In this work, we decouple global temporal representation learning from spatial refinement and arrive at a spatiotemporal model of EEG. Extensive empirical evaluations reveal that temporal conditioning improves source imaging performance relative to the unconditioned refiners, including noisy and multi-source conditions. This indicates that when instantaneous sensor topographies are ambiguous, global temporal context provides complementary information that helps to stabilize the biophysical inversion. Projecting the temporal condition into source space via $\mathbf{L}^\dagger$ prior to concatenation retains an anatomically indexed conditioning signal. While the temporal conditioning also improves the convolutional 3D-PIUNet backbone, our proposed \mymethod{} transformer supports both volumetric and surface-based sources, providing extra flexibility.

The head-model experiments identify training-operator diversity as a second important component of robust reconstruction. Mixed-pool training reduces the loss associated with applying a single-dataset specialist to another geometry pool, but broader geometry coverage alone does not eliminate sensitivity to the assumed inverse. Operator mixing addresses this additional mismatch by exposing the model to differences between the data-generating and inversion leadfields during training. Consequently, template-based reconstruction approaches the accuracy of subject-specific inversion on the evaluated held-out heads, while reconstruction with subject-specific inverses also improves. This suggests that synthetic training should represent not only the expected head geometries, but also uncertainty in the operator available at application time. These findings reduce dependence on a subject-specific inverse within the tested settings, without establishing unrestricted generalization to arbitrary montages or leadfields.

The ten-participant LEMON holdout evaluates a complementary aspect of utility. Source-power features from a model trained exclusively on synthetic EEG support age-group discrimination using a simple classifier trained on real participants. This provides evidence of downstream usefulness, rather than a general claim of classification superiority. The principal objective of this framework is not to replace task-specific sensor-space classifiers, but to support the analysis of spatially resolved activity over time. Temporal conditioning should reduce spurious reconstruction fluctuations while preserving the emergence, modulation, and disappearance of sources.

While our proposed \mymethodtc{} shines in diverse synthetic settings, the gap between simulation and biological reality remains the primary limitation. The simulator uses simplified Gaussian patches, band-limited time courses, and additive noise. Future steps of this framework could incorporate more biologically realistic priors, such as non-linear neural mass models \cite{sun2022deepsif}, structured physiological noise, and source covariances derived from resting-state fMRI connectivity \cite{Lei2011fMRIFunctionalNetworksEEGSourceImaging}. Nevertheless, increasing simulation complexity does not by itself establish biological validity. A temporally coherent reconstruction can still be biased when the underlying activity differs from the learned priors. Further evaluation should therefore distinguish noise suppression from attenuation of brief or rapidly changing sources, and assess reproducibility across recordings rather than treating temporal smoothness as evidence of accuracy.

In conclusion, decoupled temporal conditioning improves source reconstruction with a modest incremental inference cost over the corresponding unconditioned refiner in our measurements. Together with operator-mismatch training, it provides a framework for combining temporal context with physics-informed spatial estimation. Its contribution extends beyond downstream prediction toward investigating spatially resolved source dynamics within relevant time windows.
\FloatBarrier
\bibliographystyle{IEEEtran}
\bibliography{bibfile}

\end{document}